\documentclass[12pt]{article}
\usepackage[applemac]{inputenc}
\usepackage{mathptmx}
\usepackage[T1]{fontenc}
\usepackage{graphicx}
\usepackage{latexsym}
\usepackage{amsmath}
\usepackage{boxedminipage}
\usepackage{amsthm}
\usepackage{hyperref}
\usepackage{rotating}
\def\whiteghost#1{{\setbox1=\hbox{#1}\hbox to
\wd1{\vrule width 0dd depth
\dp1 height \ht1 \hfil}}}

\newcommand{\inputfigurewidth}[2]{
\begin{center}
\includegraphics[width=#1]{#2}
\end{center}}

{\theoremstyle{remark}}

\def\texturl#1{{\small\url{#1}}}
\def\footnoteurl#1{{\scriptsize\url{#1}}}
\def\titleurl#1{{\normalsize\url{#1}}}
\def\titlemail#1{{\normalsize\href{mailto:#1}{\tt #1}}}

\def\scriptLang{\textsc{CCRScript}}
\begin{document}

\title{{\textbf{A simple script language for choreography of multiple, synchronizing non-anthropomorphic
robots}\\
\normalsize --- Working draft --- }}

\author{Henning Christiansen\\
{\normalsize Roskilde University}\\
\titleurl{http://www.ruc.dk/~henning/}, \titlemail{henning@ruc.dk}}

\maketitle

\begin{abstract}
\noindent The scripting language described in this document is (in the first place) intended to be
used on robots developed by Anja Mølle Lindelof and Henning Christiansen
as part of a research project about robots performing on stage.

The target robots are expected to appear as familiar domestic objects that take their own life, so to speak,
and perhaps perform together with human players, creating at illusion of a communication between them.
In the current version, these robots' common behaviour is determined uniquely by a script
written in the language described here -- the only possible autonomy for the robots is action to correct
dynamically for inaccuracies that arise during a performance.

The present work is preliminary and has not been compared to properly to other research work in this area,
and the testing is still limited. A first implementation on small Lego Mindstorms based robots is under development
by Mads Saustrup Fox as part of his master thesis work~\cite{fox2018}.
\end{abstract}
\newpage
\tableofcontents
\newpage
\section{Introduction}
This document is intended as a reference for  \scriptLang, which is a scripting language for
robots to be used in dance, theatre and other performances.
The name of the language is an abbreviation for something like Choreographing Constrained Robots Script Language.
A script in   \scriptLang\ describes sequences of fixed  movement for one or more robots with possible synchronization when there are more
than one robot.
The robots are not assumed to have any abilities for interaction and they are expected to execute the script
faithfully with no independent decision making along the way.
The target robots are expected to appear as familiar domestic objects which has no
anthropomorphic ways of gesturing or communication such as facial expressions, arm waving, gazes etc.
They are expected to perform through elementary movements only.
Such robots and their potential have been characterized and analyzed in
a very inspiring scientific article by Bianchini et al~\cite{DBLP:conf/amf/BianchiniLMQZ14}, which we recommend as compulsory reading for anyone 
who wants to work in theory or practice with such robots.

The detailed design of   \scriptLang\ is still under development and its testing is limited, so refinements and extensions
in near future can be expected. A detailed comparison with related work in robotics and dramaturgy is still
lacking.

\section{Purpose and rationales}

\subsection{Intended users}
The expected users of the \scriptLang\ language are performance directors in a fairly wide sense, which include
also students in performance arts.
This means that using the language should not require any deep background in
programming.
Scripts in the language must be understandable and usable for the performance director in such a way that he or she can concentrate on the dramaturgic aspects.

However, the use of robots  means that a technician is likely involved
in some way or another, so technical expertise will be ready at hand, which may be helpful, for example,
when using programming language constructs to define new, high-level instructions (as explained in section~\ref{sec:user-defined-instructions}).
Possible situations of use include that the director, the technician
and perhaps performers (dancers, actors, ...) develop the script together.

The language design aims at combining simplicity with the necessary expressive
power. Instructions take as few parameters as possible, and in cases where a more detailed
control
may be relevant, this is done through optional parameters.
A sequence of movements is described in a straightforward way,
and -- if needed -- a script can be fine-tuned with additional speed and acceleration annotation.
The language is made extensible by embedding it in a standard programming language, so that new and perhaps complex patterns of movements can be added by procedural abstraction.

For simplicity, the language is almost freed from explicit timing, and only
one syncronization primitive is included, expected to be sufficiently general as to handle all
interesting cases.

\subsection{Purpose of the language}
In this version of the language, no interaction or dynamic decisions
are anticipated.
A script describes fixed sequences of movements for one or more robots,
and when once the performance is going on on the stage, there are no changes.
However, as argued elsewhere~\cite{ChristiansenLindelof2018}, the overall dramaturgic effect may appear as
a close dialogue and interaction between human and robot performers.
The actual robot implementation should include
mechanisms that correct for small inaccuracies that arise when the robots are moving,
due to
sliding, wheelspin, imprecision in the robot steering, etc.

(It is planned for future versions of \scriptLang\ to experiment
with different ways of specifying limited forms of interactions.)

\subsection{Assumptions about the robots}
\scriptLang\ is not intended for humanoid or other robots with advanced capabilities
in
body or facial expression.
They are basically objects that can move around, for example, appearing as familiar domestic 
objects, whose scenic performance and apparent liveness are created
by movements and perhaps interplay with human performers.

As indicated above, the robots are not expected to be very intelligent, technically speaking.
They should take no decisions or make any planning, except
what is needed for to produce a faithful execution of the given script.
Exceptional situations such as collisions are not considered, the robot scripts and entire
performance setup
are expected to be ``debugged'' for such before a performance.

A given robot is assumed to have some physical limitations in the movements it can perform.
These may be imposed by the actual mounting of wheels (or similar), but may also
be added by design as to produce a certain gestalt.
For example, if the robot appears as a car, it may be limited to move as a car moves,
even if wheels that are invisible to the audience allow it to move in all directions.
These details are not specified in  \scriptLang, but an implementation should include an algorithm
that generates a path from one pose\footnote{A ``pose''
in standard robot terms means the position plus angle of its forward direction
in some moment of time.} to another, that provides a ``natural way'' of moving
for the given type(s) of robots.
The currently used robots can move forwards and backwards in two dimensions, and only turn left or right within a certain
limited angle.
So, for example, a complex pattern of movements may be needed to
have a robot end up in its current positions, but turned 180 degrees.
In such situations, the different choices of how the result is obtained may
have dramaturgically very different effects and should be tailored by the director.
This may be done inserting intermediate poses and use a combination of a forward and a backward movements.

\subsection{Simulator}
In addition to the implementation on robots, \scriptLang\ is also intended to be
supported by a simulator.
The simulator should produce the same movements as the physical robots, 
and the two implementations must agree to such an extent that reliable time measurements can be made using the simulator.

It is expected that a script for a given performance is developed by help of the simulator --
and in  a combination of rehearsals on the stage (perhaps together with humans performers).
The simulator can be used for ``debugging'' the script, so no undesired situations, such as collisions, are likely to occur, and for testing the overall pattern of actions performed by the robots.

\section{The scripting language}
The language is independent of a particular implementation, but is intended to be used for both controlling
actual robots in physical space and for produce animations in a simulator.
These two implementations should agree to the extent that in case the robots are not subject to unexpected
incidents, the observable patterns of movements should appear is identical.

The robot software should make corrections for small accuracies that arise due to
sliding, wheelspin and similar phenomena, so that the robot always ends up in at specified pose (perhaps
slightly later that expected).  There is no absolute timing that the robots need to keep up to, the principle
is that ``things take the time they will take'' (this is a design decision imposed in order to simplify \scriptLang), determined
by possible speed and acceleration parameters (e.g., default values) and the physical conditions.
The simulator and robot implementation must agree to such an extent that reliable time measurements can be made
using the simulator.

The detailed movements generated by an instruction depend on the robots' the physical constraints;
this is not specified in  \scriptLang, but an implementation should include an algorithm
that generates a path that provides a ``natural way'' of moving
for a robot with the given gestalt and its physical limitations.

The simulator and the robot implementation should both produce the same movements, and the simulator
may give diagnostic messages in case the algorithm used cannot create a ``reasonable'' path from one place to another
-- or if the path passes through illegal positions (as defined below).

The only way that a director can modify a  path from one pose to another is to insert intermediate poses and
perhaps switch between forward and backward movements (analogous to parking a car in a narrow space).

\subsection{Embedding in a general purpose programming language}
\scriptLang\ should be embedded a standard programming language in order to provide:
\begin{itemize}
 \item Ways of naming different robots and referring to them; typically by program variables.
  \item Standard datatypes such as text strings and numbers and abilities to do standard arithmetic and such.
  \item Procedural abstraction that makes it possible to put together sequences of instructions and give them a new name, thus
  extending \scriptLang\ with new instructions; see examples in section~\ref{sec:user-defined-instructions}, below.
\end{itemize}
The current prototype simulator is embedded in Processing that follows Java syntax -- which (for the part used here) is
quite mainstream, being
more or less identical to C, C++, C\#, etc.

Among the standard datatypes, we assume colours; the actual syntax may depend on the embedding programming language.
Processing uses RGB colours, so, for example,
\texttt{color(255,128,128)} indicates a pink colour.

\subsection{Coordinate system, poses and time}
The robots move within a scene which is assumed to be rectangular.
The $x$-axis is parallel to the border of the scene towards the audience, and the $y$-axis points into the scene as seen from the audience.
The point $\langle 0,0\rangle$ is placed middle front in the scene, such that points to the right of the middle has positive $x$ values
and points to the left negative values. The $y$-values are zero at the front of the scene and grow positively into the scene; see illustration below.

The basic unit is one meter, and other units are available; explained below.
The overall dimension of the scene should be defined before any other \scriptLang\ instructions are use.
It may, for example, be done as follows.
\begin{verbatim}
float sceneWidth = 10;
float sceneDepth = 5; 
\end{verbatim}
This defines a  scene which is 10 meters wide and 5 meters deep; thus $x$-coordinates within the scene ranges between
$-5$ and $5$ and $y$-coordinates  between
$0$ and $5$;
any position outside this range is called \emph{illegal}, and it is also possible to define illegal areas within the scene (see section~\ref{sec:instructions:illegal:areas}).
There are no mechanisms in the language that prevent a robot moving to an illegal position.\footnote{The director
is expected to ``debug'' the script using a simulator that will issue suitable warnings.
Robots in illegal position can be relevant, e.g., if a robot enters the scene by a door or exits the same way.}

A number without explicit units used for a position or distance is interpreted as meters; the following named units
are also available:
\begin{itemize}
  \item \texttt{sw} referring to the width of the scene.
  \item  \texttt{hsw} referring to half the width of the scene.
  \item  \texttt{sd} referring to the depth of the scene.
   \item  \texttt{m} for meters (it is redundant, but included for completeness).
\end{itemize}
Notice: As  \scriptLang\ is embedded in a standard programming language syntax,
a multiplication operator must be use when using these units; for example, the quarter of the width of the scene may be indicated as.
The coordinate system and coordinates of selected points are shown as follows.
\inputfigurewidth{0.8\textwidth}{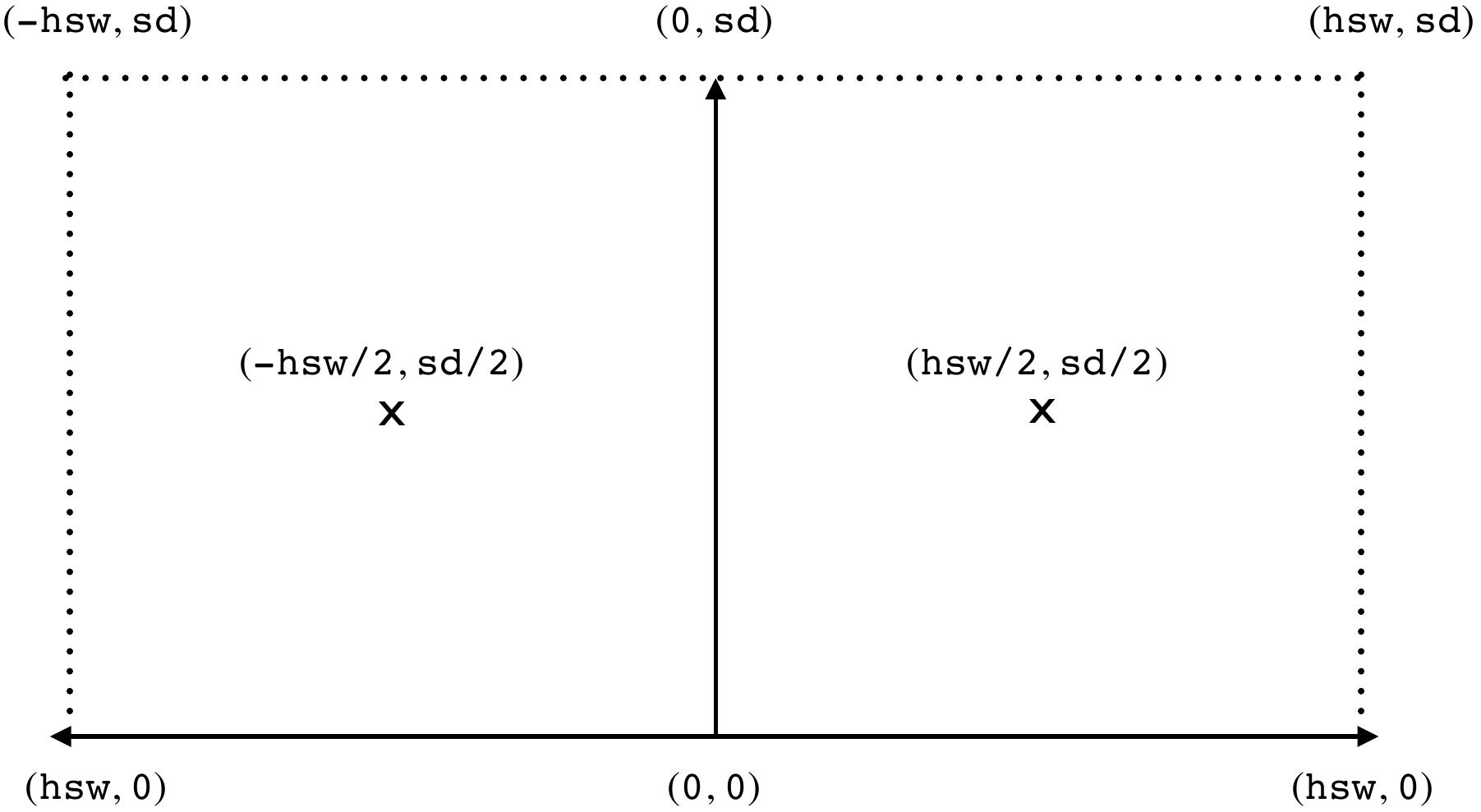}
Angles are indicated in degrees, and the following names can be used for absolute directions;
the direction of North points into the scene.
\begin{itemize}
  \item \texttt{north} for the direction into the scene, synonymous with \texttt{0} degrees.
  \item  \texttt{east},  \texttt{south},  \texttt{west}, \texttt{northEast},  \texttt{southEast}, \texttt{northWest}, \texttt{southWest} analogously.
  \item  \texttt{nne} for North-North-East and analogously \texttt{ene}, \texttt{ese}, \texttt{sse}, \texttt{ssw}, \texttt{wsw},  \texttt{wnw}, \texttt{nnw}.
\end{itemize}
Arithmetic can be used, so, e.g., one degree starboard of east can be indicated as \texttt{east+1} or as \texttt{91}.\footnote{Notice that such compass angles are different from a traditional mathematical way of considering angles;
a mathematician will typical indicate an angle of zero degrees pointing to what a compasicist will
call east (i.e., $+90^\circ$). And when a mathematician says an angle is \emph{in}creases, the
compasicist will correct him and say its is \emph{de}creasing.}

At any given time, a robot has a \emph{pose}, which is the combination of $x$ and $y$ coordinates and a direction (assumed to be the direction it is heading, or, metaphorically, the direction it is ``looking''). So, for example, a robot standing at the very centre of the scene, looking towards the audience
has the pose $\langle \texttt{0}, \texttt{0.5*sd}, \texttt{south}\rangle$.
If needed, poses may be treated as data objects, but this is actually not necessary for writing scripts.
It the programming syntax, a special datatype is used together with a function to construct a pose as a data object:
\begin{verbatim}
    pose(0, 0.5*sd, south)
\end{verbatim}
\emph{Time} is measured in seconds, but the use of explicit timing
in  \scriptLang\ is reduced to a minimum: the specified robot movements take the time they will take;
time is used explicitly for wait instructions, e.g., when instructing a robot to stand still for 10 seconds before it goes on with the next
given movement.

\subsection{Instructions}

\subsubsection{Introducing new robots}
In this version of \scriptLang, the physical properties of the robots are not considered explicitly,
but they can be specified by a few attributes that may be used by a simulator.
The syntax used in the Processing version of \scriptLang, is illustrated as follows.
\begin{verbatim}
    Robot nille = robot("Nille", color(255,128,128));
    Robot frederik = robot("Frederik", color(128,128,255));
\end{verbatim}
Two robots are defined; they can be referred to later in the script through the program variables
\texttt{nille} and \texttt{frederik}; the parameters to the robot introdiction function \texttt{robot($\cdots$)}
indicates a name (as a text string) and a colour that may be used by a simulator. (The sample colours are pink and light-blue.) 

Each robot need to be given an initial pose, i.e., so to speak be placed into the scene, before movement commands 
give sense. The two robots defined above may be placed as follows.
\begin{verbatim}
    initialPose(nille,    hsw/2+1, sd/4, west);
    initialPose(frederik, hsw/2, sd/4+1, south);
\end{verbatim}
This positions are shown in this figure; the heading direction is indicated by the arrow;  robot \texttt{nille} is  pink, \texttt{frederik} light blue.
\inputfigurewidth{0.8\textwidth}{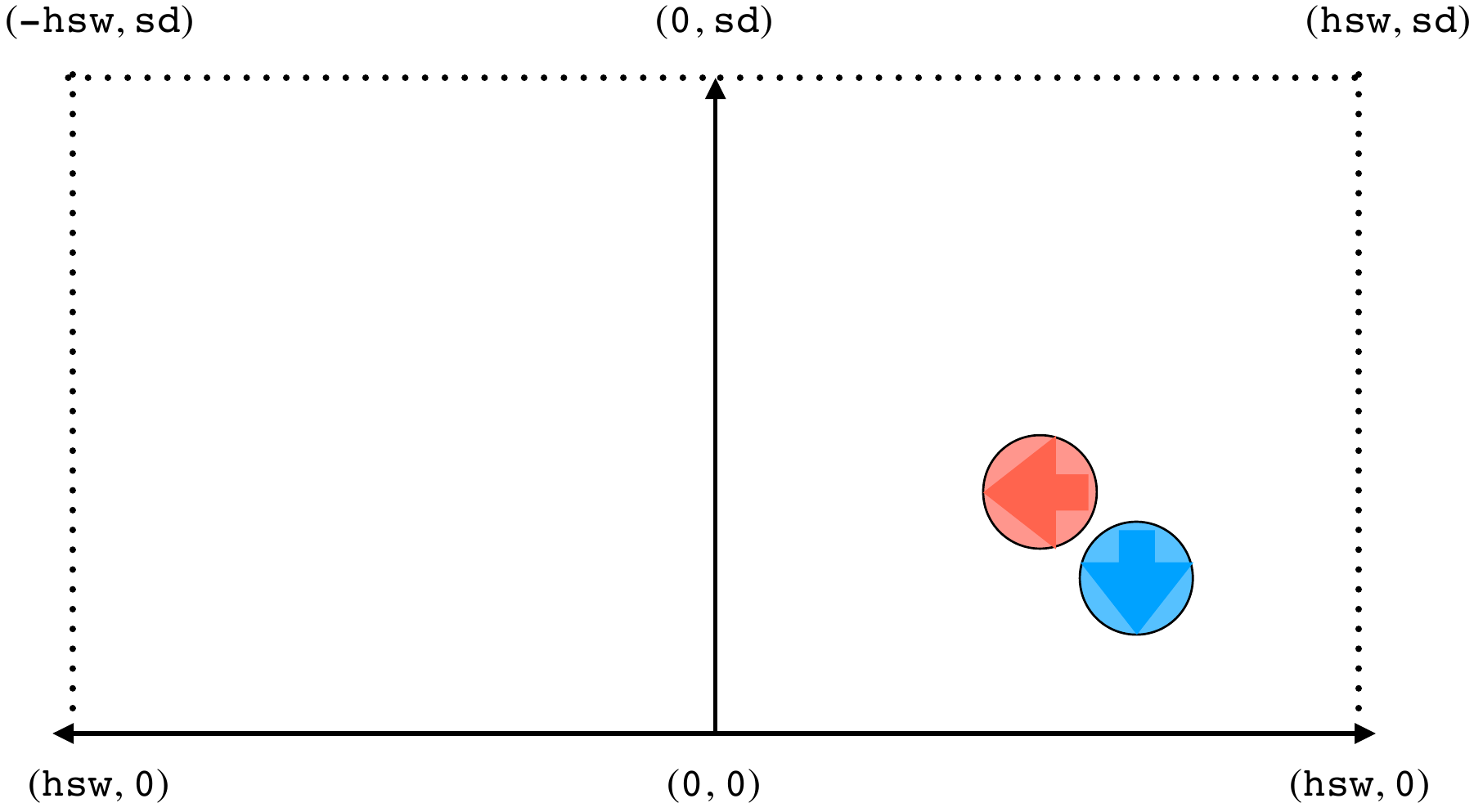}
The use of poses as data objects is shown in the following code fragment (that does the same as the above).
\begin{verbatim}
    Pose p1 = pose(hsw/2+1, sd/4, west);
    Pose p2 = pose(hsw/2, sd/4+1, south);
    initialPose(nille,    p1);
    initialPose(frederik, p2);
\end{verbatim}
It is not enforced that the robots must by placed physically in those positions indicated in the script.
It is, of course, recommended to do so; or at least approximatively as a good implementation
should correct for inaccuracies.

\subsubsection{Move instructions}
Some instructions tell the robot to move to a new, given pose from whatever pose it is at when it starts
executing the instruction.
For example: 
\begin{verbatim}
    moveTo(nille, -hsw/2, sd/2, south);
\end{verbatim}
Unless otherwise specified, the robot \texttt{nille} is assumed to be at a standstill, it will accelerate
to a normal speed, follow the generated path in a forward direction, and finally decelerate and come to a standstill at the indicated pose.
Alternatively, the movement can be done having the robot moving backwards:
\begin{verbatim}
    moveToBacking(nille, -hsw/2, sd/2, south);
\end{verbatim}
The following, alternative syntax may apply when poses are considers as data objects.
\begin{verbatim}
    moveTo(nille, pose(-hsw/2, sd/2, south));
    moveToBacking(nille, pose(-hsw/2, sd/2, south));
\end{verbatim}
The remaining instructions specify relative movements.
To move forward or backward in the direction indicated by the current pose
is exemplified as follows:
\begin{verbatim}
    move(nille, 2);
    moveBacking(frederik, 2);
\end{verbatim}
Robot \texttt{nille} moves two meters forward according to its current pose,  \texttt{frederik} moves two meters
backwards,

There are instruction for moving along a segment of a circle
\begin{verbatim}
    circleRight(nille, 1, 90);
\end{verbatim}
It means: The robot \texttt{nille} should turn to the right by traversing a circle with radius \texttt{1} meter and an angle of \texttt{90} degrees.
There are similar instructions for circling left an circling backwards. Examples:
\begin{verbatim}
    circleRight(nille, 1, 90);
    circleLeft(nille, 1, 90);
    circleRightBacking(nille, 1, 90);
    circleLeftBacking(nille, 1, 90);
\end{verbatim}
The angle can be arbitrarily high, so for example the following instruction specifies that
\texttt{nille} rotates 100 times without getting anywhere.
\begin{verbatim}
    circleRight(nille, 1, 36000);
\end{verbatim}
The relative move instructions should executed such that the robot ends up at the pose, where they would end if all previous
move instructions had been executed with mathematical precision, i.e., the execution of the instructions
on the physical robots should adjust for errors in the pose they start from.

\subsubsection{Wait instruction}
A robot can be told to wait for a given number of seconds, for example:
\begin{verbatim}
    wait(nille, 2);
\end{verbatim}

\subsubsection{Independent execution for each robot}
Each robot executes its own instructions independently of other robots (unless when synchronization is used;
section~\ref{sec:synchronization}, below).
This means that two sequences of instructions for two different robots
can be intertwined in different ways without changing the overall behaviour.
Consider as an example the following \scriptLang\ fragment.

\begin{verbatim}
    initialPose(nille, hsw/2, 0, north);
    initialPose(frederik, -hsw/2, 0, north);
    moveTo(nille, hsw/2, 3, north);
    moveTo(frederik, -hsw/2, 3, north);
    moveTo(nille, hsw/2+2, 3, east);
    moveTo(frederik, -hsw/2+2, 3, east);
\end{verbatim}
The two robots will start moving simultaneously and continue moving in parallel.
We may, e.g., group the instructions for each robot as follows without changing the meaning.
\begin{verbatim}
    initialPose(nille,    hsw/2, 0, north);
    moveTo(nille, hsw/2, 3, north);
    moveTo(nille, hsw/2+2, 3, east);
    initialPose(frederik, -hsw/2, 0, north);
    moveTo(frederik, -hsw/2, 3, north);
    moveTo(frederik, -hsw/2+2, 3, east);
\end{verbatim}

\subsubsection{Fine-tuning movements -- EXPERIMENTAL VERSION}
The robots are assumed to have a standard maximum speed and standard acceleration and deceleration.
Unless otherwise specified, these will be used when interpreting the move instructions.
Speed is measured in meters per second and acceleration / deceleration in meters per second$^2$.

These figures can be set for all robots at the same time or each robot individually. Examples:
\begin{verbatim}
    maxSpeed(2);
    acceleration(frederik, 0.5);
    deceleration(nille, 0.2);
\end{verbatim}
Instead of an absolute numbers, the following constants can be used.
\begin{itemize}
  \item [\emph{max}:] the maximum possible value, determined by the physical robot.
  \item  [\emph{std}:] the standard value for this robot.
\end{itemize} 
Dynamic changes in speed during a move instruction may specified by a text string that can be added
as an extra argument to move instruction.
The following characters have a special meaning inside these instructions.
\begin{itemize}
  \item [\texttt{!}:] if appearing as the first character in such a string, it means to start the movement
  as fast as possible; at the end of a string, it means brake as late and as hard as possible in order to stop at the specified pose;
  in any other position, it means break as hard as possible and immediately start as fast as possible (likely experienced as a sort of jump).
  \item [\texttt{=}:] if appearing as the first character in such a string, it means to continue
  with the speed from the previous instruction; at the end of a string, it means suppress deceleration when the specified pose is reached.
  \item [\texttt{+}:] increase speed.
  \item [\texttt{-}:] decrease speed.
\end{itemize}
Any other character means continue as you are doing.
Control strings consisting of a single ``\texttt{!}'', resp.\ ``\texttt{=}'', are interpreted as ``\texttt{!!}'', resp.\ ``\texttt{==}''.
The use of ``\texttt{=}'' works best when an instruction whose control string ends with ``\texttt{=}''
is followed by another instruction whose control string starts with ``\texttt{=}'' (This is not enforced,
and if used differently, the implementation may just ignore the ``\texttt{=}''.

The ``\texttt{=}'' character in control strings can be used to produce a `smooth concatenation
of different move instructions.
Example:
\begin{verbatim}
    initialPose(nille,    hsw/2, 0, north);
    moveTo(nille, hsw/2, 3, north, "=");
    circleRight(nille, 1, 90, "=");
    moveTo(nille, hsw/2+2, 4, east, "=");
\end{verbatim}
The length of a string is not fixed, and the implementation will try to
match each character with an equal time interval during the movement.
Unless overridden by ``\texttt{!}'', a normal acceleration / deceleration and the star/end is assumed.
Examples:
\begin{verbatim}
    "++__++____----"
    "!!!!!!!!!!"
\end{verbatim}
The first one means: accelerate  normally, continue increasing the speed (\texttt{++}); continue a bit with the obtained speed  (\texttt{\_\_});
accelerate more (\texttt{++}); continue with obtained speed (\texttt{\_\_\_\_}), and then gradually decreasing the speed to normal and
the stop by a normal deceleration.
The second one is like interpreted as a very little gracious way of jumping along.

The following example shows the syntax for using such control strings.
\begin{verbatim}
    moveTo(nille, -hsw/2, sd/2, south,  "++__++____---!");
\end{verbatim}

\subsubsection{Synchronization}\label{sec:synchronization}
There is only one instruction for synchronization, and it implements the rule ``everyone waits for the slowest one''.
\begin{verbatim}
    synchronize();
\end{verbatim}
This means that each robot finishes all instructions up to the synchronization instruction and stops;
when all robots have finished and stopped this way, they all proceed to their next sequence of instructions.

There are variants of the instructions that allow to specify which robots should synchronize. This may be useful
when there are three or more robots in action. Example:
\begin{verbatim}
    synchronize(nille,frederik);
\end{verbatim}
Robots \texttt{nille} and \texttt{frederik} synchronize while all other robots continue whatever they are doing.

\subsection{User-defined instructions by means of abstraction in the embedding programming language}\label{sec:user-defined-instructions}
The following is an example of how a new instruction be defined using a function definition
in Processing and Java.
\begin{verbatim}
    void steps(Robot rob, int n) {
      for(int i=1; i<=n; i++) {move(rob, 0.3); moveBacking(rob, 0.3);}
    }
\end{verbatim}
The new instruction can be used together with other instructions, exemplified as follows.
\begin{verbatim}
    moveTo(nille, 0, 1, south);
    steps(nille, 7); 
    moveToBacking(nille, 0, sd-1, south, "+++++!"),
\end{verbatim}
Robot \texttt{nille} will go to the centre of the scene, one meter from the edge of the scene, looking at the audience.
Then it steps back and forth 30 cm seven times, appearing to be undetermined, and then it flees very fast to the back of the scene,
and brakes hard one meter from the back cloth.

Here is another definition of a new instruction parameterized by two robots and a position.
When called for two robots and a given position, the robots will first find their way to the left, resp.\ right, of the indicated position;
when both have arrived, they perform a ritual in which they mirror each other.
\begin{verbatim}
    void meetAndGreat(Robot r1, Robot r2, float x, float y) {
      moveTo(r1, x-1, y, east); 
      moveTo(r2, x+1, y, west);
      synchronize(r1,r2);
      moveTo(r1, x-0.25, y, east);
      moveTo(r2, x+0.25, y, west);
      wait(r1,1); wait(r2,1);
      moveToBacking(r1, x-1, y, east); 
      moveToBacking(r2, x+1, y, west);
      wait(r1,0.5); wait(r2,0.5);
    }
\end{verbatim}
It may be uses, e.g., as follows in a script.
\begin{verbatim}
    meetAndGreat(nille, frederik, float x, float y);
\end{verbatim}

\subsection{Defining forbidden areas in the scene, references points and grids}\label{sec:instructions:illegal:areas}
These instructions affect only execution in a simulator.
A forbidden area is a rectangle which will be shown with a special colour; the robot
is not prevented from entering these areas, but the simulator will issue a warning.
Reference points and grids will be drawn  by the simulator in the depiction of the scene, but serves
no other purposes.
In the current version of \scriptLang, the settings made by these instructions are final: once set, they cannot be un-done.
The following fragment a script shows the current options. 

\begin{verbatim}
    grid();
    referencePoint(color(255,0,0),hsw/2,sd/2);
    referencePoint(color(0,255,0),-hsw/2,sd/2);
    forbiddenArea("green stuff", color(192,255,192),
                                 -hsw+0.5, sd-1.75,   -hsw+3, sd-2 );
    forbiddenArea("purple stuff", color(192,192,255),
                                 0.5,0.5,   hsw-2, 1) );
\end{verbatim}
The yet uninhabited scene may be shown as follows, assuming a width of 10 meters and a depth of 5 meters.
\inputfigurewidth{0.8\textwidth}{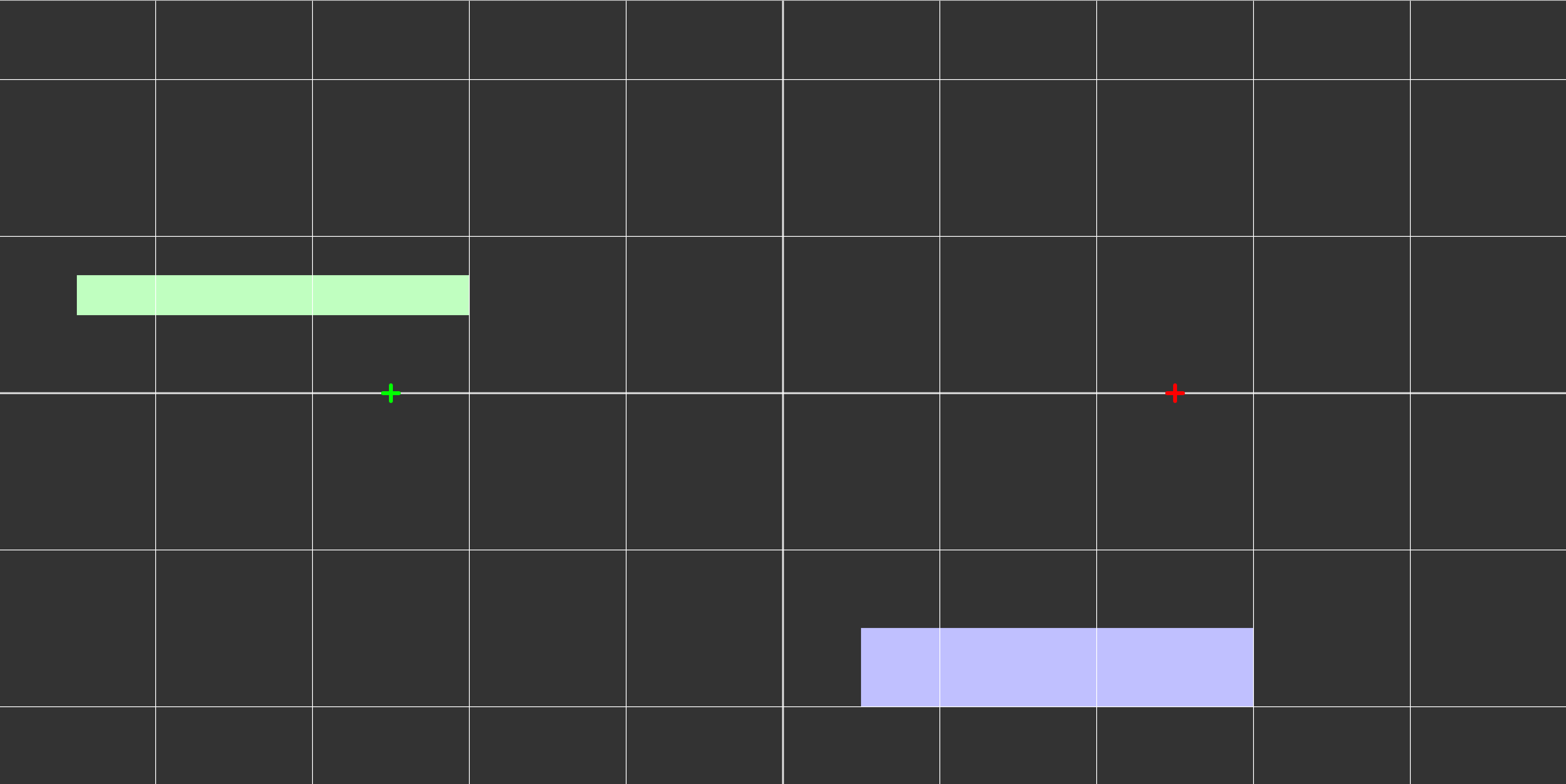}


\subsection{Possible extension considered for future versions}
\subsubsection{Different types of robots}
Relevant if we have robots with different gestalt and-or physical limitations,
which may mean also that the implementations (on the robots and in the simulator) may have different path finding algorithms.
The type of each robot should be indicated when the robot is introduced.
\begin{verbatim}
    Robot nille = robot("Nille", type1, color(255,128,128));
    Robot frederik = robot("Frederik", type2, color(128,128,255));
\end{verbatim}

\subsubsection{Different path-finding algorithms}
As we have tested different algorithms for find a »best« path from one pose to another,
we have observed that each have its advantage and may give a different dramaturgic result.
An extra argument may be added to the \texttt{moveTo} command as shown in the following.
For example: 
\begin{verbatim}
    moveTo(nille, -hsw/2, sd/2, south,  forward);
    moveTo(nille, -hsw/2, sd/2, south,  backward);
    moveTo(nille, -hsw/2, sd/2, south,  smart);
\end{verbatim}
While the two first ones are synonymous with \texttt{moveTo} and  \texttt{moveToBacking} commands that are
shown above, the \texttt{smart} option may allow the algorithm to combine forward and backward movements,
which can give shorter and more »natural« paths that reduces the amount of circling.

\subsubsection{Including human performers' behaviour}
It may be interesting in the simulator also to include the movements of human performers.
The humans will in this respect appear as just another type of robot, the only difference is that
they are ``controlled'' only in the simulator.

\subsubsection{Other robot abilities}
In case the robots are equipped with blinking lamps, loudspeakers or other sound devices, waving flags, etc.,
\scriptLang\ should be extended with instructions to use those.
This will require a bit of design thinking in case these actions can be performed while the robot
is moving (i.e., the same robot is performing two or more actions at the same time).

\subsubsection{Instructions for specifying interaction and dynamic
adaptation}
As described above, the robots do not react to the actions of a human performer, and the only way to create such an illusion is that the
human performer knows exactly what the robot will be doing and acts accordingly.
As described elsewhere this can be done in a very convincing way, but requires top-professional human performers~\cite{}.

It may be interesting to extend the language with facilities that allows an interaction, limited in time and space.
For example, the robot may be instruction to stay on a certain line segment and in a given period maintain a constant
distance to the object in front of it. If this object is a human dancer, that takes a jump towards the robot, the robot
should perform a similar jump backwards.

These ideas are currently only at a speculative level, and more experience with  robots controlled via \scriptLang\ before
such facilities are considered,

\section{A prototype implementation of the simulator}\label{sec:prototype}
A prototype of a simulator has been implemented in Processing.
A script is supposed to be included as the body of a function called \texttt{manuscript()}.
The implementation includes an algorithm that can calculate a ``natural'' path between
two poses, which fits the movements of a robot which can only turn left or right within a certain limited angle.
Such a path is described as a sequence of small steps, and an additional translation
into motor control commands is needed in order to use this algorithm for the physical robots.
An alternative algorithm that always can find a way from on pose to another, consisting of at most
two circle segments and a straight line segment has also been tested.


\section{Guidelines for a fully developed user interface for the simulator}
The following facilities are desirable in a final  version of a simulator to be used in performance production.
\begin{itemize}
\item A straightforward transfer of a script to a robot implementation.
\item Ability to run simulator and robots concurrently (as to see if the robots actually do what they are supposed to).
\item Starting and stopping the simulation, normal and reverse simulation, slow and fast simulation.
\item Automatic annotation of the given script with timing information,
\item Changing viewpoint, between seeing things from above and perspective views, as soon from the audience
\end{itemize}
It has not been considered whether it is interesting to use high quality, 3D graphics (e.g., using Unity 3D),
and if arbitrary viewpoints are useful.
Perhaps a schematic view will still give the best overview.

\bibliographystyle{plain}
\bibliography{robots}
\end{document}